\documentclass[10pt,letterpaper,twocolumn]{article}

\usepackage[margin=1.9cm]{geometry}

\usepackage[hyphens]{url}
\usepackage{breakurl}
\newenvironment{packed_itemize}{
\begin{itemize}
  \setlength{\itemsep}{1pt}
  \setlength{\parskip}{0pt}
  \setlength{\parsep}{0pt}
}{\end{itemize}}

\begin{document}

\twocolumn[
\title{SysML'19 demo: customizable and reusable Collective Knowledge pipelines to automate and reproduce machine learning experiments}

\author{Grigori Fursin \\
\\
cTuning foundation; dividiti\\
\\
Demo website: \url{https://github.com/ctuning/ck/wiki/Demo-SysML19}\\
}

\date{February, 2019}

\maketitle

%%%%%%%%%%%%%%%%%%%%%%%%%%%%%%%%%%%%%%%%%%%%%%%%%%%%%%%%%%%%%%%%%%%%%%%%%%%%%%%%%
\begin{abstract}

Reproducing, comparing and reusing results from machine learning 
and systems papers is a very tedious, ad hoc and time-consuming process.
I will demonstrate how to automate this process 
using open-source, portable, customizable and CLI-based 
Collective Knowledge workflows and pipelines
developed by the community~\cite{CK}.
I will help participants run several real-world non-virtualized CK workflows
from the SysML'19 conference, companies (General Motors, Arm)
and MLPerf benchmark~\cite{MLP} to automate benchmarking and co-design of efficient 
software/hardware stacks for machine learning workloads.
I hope that our approach will help authors reduce their effort 
when sharing reusable and extensible research artifacts 
while enabling artifact evaluators to automatically 
validate experimental results from published papers 
in a standard and portable way.

\end{abstract}

\vspace{0.3cm}


\vspace{0.3cm}

]

%%%%%%%%%%%%%%%%%%%%%%%%%%%%%%%%%%%%%%%%%%%%%%%%%%%%%%%%%%%%%%%%%%%%%%

\section{Introduction}

Ten years ago I started collaborating with the community to automate
and reproduce complex systems and machine learning experiments~\cite{Fur2009}.
This experience helped us introduce and standardize validation of experimental results 
from published papers at PPoPP, CGO, Supercomputing and other systems conferences 
during the so-called Artifact Evaluation process~\cite{AE}.

Voluntarily AE typically works as follows. 
Once a paper is accepted, the authors can submit a related
artifact (code, data, models, results) to a special committee usually 
formed from senior graduate students and postdoctoral researchers. 
This committee attempts to reproduce results using
submitted artifacts based on the standard ACM procedure 
we contributed to a few years ago~\cite{ACM}: 
check if the artifact is permanently archived, 
follow its installation instructions and verify functional correctness,
partially replicate results from the paper, assign
reproducibility badges and so on.

The good news is that the number of papers
participating in AE has dramatically increased over the years.
For example, PPoPP'19 has set a new
record with 20 papers passed through AE out of 29 papers
accepted. 
Supercomputing'19 has even made artifact description appendices
mandatory for all submitted papers. 
ACM also provided advanced article search based
on reproducibility badges in the ACM Digital Library.

The bad news is that the lack of a common experimental framework, 
common research methodology, common formats and stable APIs 
places an increasing burden on evaluators to validate and compare 
a growing number of very complex artifacts. 
Furthermore, having too many papers with too many ad-hoc artifacts 
and Docker snapshots is almost as bad as not having any, 
since they cannot be easily reused, customized and built upon.

These problems are particularly noticeable in co-design projects 
developing efficient software and hardware stacks for emerging workloads 
including artificial intelligence and machine learning.
Such stacks must be continuously benchmarked, optimized and compared
across rapidly evolving machine learning models, data sets, frameworks, 
libraries, compilers and hardware architectures.

However, when overviewing more than 100 papers during artifact
evaluations~\cite{AE}, I noticed that many of them use very similar 
experimental setups, scripts, pipelines, benchmarks, models, 
data sets and platforms.
This motivated me to develop Collective Knowledge (CK),
an open-source, extensible and CLI-based framework with a simple Python API
to help researchers automate their repetitive tasks such as detecting software
dependencies in a native, non-virtualized environment;
installing missing packages; downloading data sets and models;
compiling and running programs, implementing autotuning and co-design pipelines;
crowdsourcing time-consuming experiments across computing
resources provided by volunteers; automatically reproducing, plotting
and validating experimental results, and so on~\cite{CK}.
Such approach helps a growing number of organizations
apply DevOps principles to their research while integrating
shared CK pipelines and components with their existing 
projects~\cite{USE}.

%%%%%%%%%%%%%%%%%%%%%%%%%%%%%%%%%%%%%%%%%%%%%%%%%%%%%%%%%%%%%%%%%%%%%%%%%%%%%%%%%
\section{Demonstration}

I will demonstrate how to prepare, run, visualize and reuse several CK pipelines 
for image classification from the 1st ACM ReQuEST tournament at ASPLOS'18                       
aiming to collaboratively co-design efficient SW/HW stacks 
for deep learning and reproduce results using the established AE methodology~\cite{REQ}:

\begin{packed_itemize}
 \item \textbf{Edge:} TFLite with MobileNets on ARM-based Android devices - this CK pipeline will be reused in MLPerf~\cite{MLP,CK-MLPERF,CK-ANDROID}
 \item \textbf{Cloud:} Intel Caffe with 8-bit ResNet-50 on AWS c5.18xlarge instances - this CK pipeline was validated by Amazon colleagues~\cite{AIC}
 \item \textbf{FPGA:} TVM/MXNet with 8-bit ResNet-18 on an FPGA simulator~\cite{RESCUE-TVM}
\end{packed_itemize}

I will also show CK pipelines from several accepted SysML'19 papers which successfully passed Artifact Evaluation~\cite{SAE}.
The participants should be able to repeat all interactive steps on their own computers
while learning CK concepts and interfaces.
It can help future SysML authors share reusable pipelines with a common API 
to automate and accelerate validation of experimental results from their articles.
The participants will need the following software to reproduce my demo~\cite{CKI}:

\begin{packed_itemize}
 \item Linux or MacOS with bash (CK also supports Windows but several third-party tools used in my demo work only on Unix)
 \item Python 2.7 or 3.3+ 
 \item Git client
 \item CK CLI 
 \item Any GCC, LLVM or ICC
 \item Internet access
 \item Optional: Android NDK and SDK to test ML pipelines on Android devices
\end{packed_itemize}

%%%%%%%%%%%%%%%%%%%%%%%%%%%%%%%%%%%%%%%%%%%%%%%%%%%%%%%%%%%%%%%%%%%%%%%%%%%%%%%%%
\section{Short biography}

I run the non-profit cTuning foundation where I work with the community
to develop a methodology and open-source tools for collaborative, reproducible 
and reusable ML and systems research.
I am also the architect of the "Collective Knowledge"
technology~\cite{CK} to automate, crowdsource and reproduce experiments.
Since 2014 I have helped to initiate and standardize Artifact Evaluation
at different systems conferences~\cite{AE} where I also introduced 
a unified Artifact Appendix.

%%%%%%%%%%%%%%%%%%%%%%%%%%%%%%%%%%%%%%%%%%%%%%%%%%%%%%%%%%%%%%%%%%%%%%%%%%%%%%%%%
\bibliographystyle{abbrv}
\bibliography{paper}

\end{document}